\documentclass[10pt,conference]{IEEEtran}
\IEEEoverridecommandlockouts
\usepackage{cite}
\usepackage{amsmath,amssymb,amsfonts}
\usepackage{algorithmic}
\usepackage{graphicx}
\usepackage{hyperref}
\usepackage{textcomp}
\usepackage{xcolor}

\usepackage{tikz-cd}
\usetikzlibrary{arrows.meta, positioning}

\tikzset{
  block/.style={
    rectangle,
    draw,
    rounded corners,
    minimum height=1cm,
    minimum width=1.6cm,
    align=center,
    font=\small
  },
  arrow/.style={
    ->,
    thick
  }
}

\def\BibTeX{{\rm B\kern-.05em{\sc i\kern-.025em b}\kern-.08em
    T\kern-.1667em\lower.7ex\hbox{E}\kern-.125emX}}
\begin{document}

\title{Kolmogorov-Arnold graph neural networks for chemically informed prediction tasks on inorganic nanomaterials}

\author{\IEEEauthorblockN{1\textsuperscript{st} Nikita Volzhin}
\IEEEauthorblockA{\textit{Department of Mathematics and Science} \\
\textit{American University in Bulgaria}\\
Blagoevgrad, Bulgaria \\
ORCID: 0009-0002-1705-0680\\
nvv220@aubg.edu}
\and
\IEEEauthorblockN{2\textsuperscript{nd} Soowhan Yoon}
\IEEEauthorblockA{\textit{Department of Mathematics and Science} \\
\textit{American University in Bulgaria}\\
Blagoevgrad, Bulgaria \\
ORCID: 0000-0002-7979-4880\\
syoon@aubg.edu}

}
\maketitle

\begin{abstract}
The recent development of Kolmogorov-Arnold Networks (KANs) has found its application in the field of Graph Neural Networks (GNNs) particularly in molecular data modeling and potential drug discovery \cite{b11}. Kolmogorov-Arnold Graph Neural Networks (KAGNNs) expand on the existing set of GNN models with KAN-based counterparts. KAGNNs have been demonstrably successful in surpassing the accuracy of MultiLayer Perceptron (MLP)-based GNNs in the task of molecular property prediction. These models were widely tested on the graph datasets consisting of organic molecules. In this study, we explore the application of KAGNNs towards inorganic nanomaterials. In 2024, a large scale inorganic nanomaterials dataset was published under the title CHILI (Chemically-Informed Large-scale Inorganic Nanomaterials Dataset), and various MLP-based GNNs have been tested on this dataset \cite{b2}. We adapt and test our own KAGNNs appropriate for eight defined tasks. Our experiments reveal that, KAGNNs frequently surpass the performance of their counterpart GNNs. Most notably, on crystal system and space group classification tasks in CHILI-3K, KAGNNs achieve the new state-of-the-art results of 99.5 percent and 96.6 percent accuracy, respectively, compared to the previous 65.7 and 73.3 percent each.
\end{abstract}

Code: \url{https://github.com/Nikitavolzhin/KAGNN-for-CHILI}\par
Datasets: \url{https://github.com/UlrikFriisJensen/CHILI}

\section{Introduction}\label{sec:intro}

In 2024, Friis-Jensen et al. published a large-scale inorganic nanomaterials dataset called CHILI (Chemically-Informed Large-scale Inorganic Nanomaterials Dataset) \cite{b2}. Seventeen machine learning tasks were posed within it: classification, regression, and structure generation problems. Wide range of graph neural network (GNN) models have been tested on it. We find three of these models, namely Graph Convolutional Network (GCN), Graph Isomorphism Network (GIN) and EdgeCNN, to have fertile grounds for a further investigation.

Kolmogorov-Arnold Network (KAN) has been introduced by Liu et al. as an alternative to the conventional multi-layer perceptron (MLP) models for machine learning \cite{b9}. This new architecture has been recently adopted in graph neural networks in various forms \cite{zhang, b30, b27, b1, b9, b11}. While Kolmogorov-Arnold Graph Neural Networks (KAGNNs) have been shown to outperform many of the competing models on several organic-molecule datasets \cite{b1, b11, b27}, they have not yet been tested on inorganic nanomaterials. We employ Kolmogorov-Arnold layers on the three aforementioned GNN architectures on CHILI dataset.

We summarize our experiment on CHILI as follows:
\begin{itemize}
  \item We propose KAEdgeCNN (see formula \ref{eq12}), an EdgeCNN based Kolmogorov-Arnold Network, and utilize the existing Kolmogorov-Arnold GCN (KAGCN) and Kolmogorov-Arnold GIN (KAGIN) models \cite{b1} for a performance comparison with EdgeCNN, GCN and GIN, respectively. 
  \item We conduct a hyperparameter search for KAGCN, KAGIN, and KAEdgeCNN on CHILI-3K, smaller of the two data collections in CHILI, and share the obtained hyperparameters for the benchmark test on both CHILI-3K and CHILI-100K.
  \item We follow the same benchmark test procedures as reported in \cite{b2} for a robust comparison.
\end{itemize}

Our experiment led to the following discoveries.
\begin{itemize} 
  \item We report a significant improvement of the performance by KAGNNs on classification tasks. Especially on crystal system and space group classifications, KAGCN and KAEdgeCNN respectively achieve under the weighted F1 score the new state-of-the-art accuracy of 99.5 percent and 96.6 percent compared to the previous 65.7 percent and 73.3 percent by EdgeCNN.
  \item It has been observed that the introduction of KAN layers diminishes the noticeable advantage EdgeCNN has over other models (e.g., GCN). In other words, the ``KAN-trick" resolves the performance gap issue between GNNs and enables various models to thrive.
\end{itemize}

The rest of this paper is organized as follows. Section~\ref{sec:background} reviews KAN and existing vanilla GNNs. Section~\ref{sec:related} provides a short overview of relevant works. In section~\ref{sec:method}, we define the KAN-based GNN architectures used for our experiment, describe carefully our experimental procedure, and report the results. We convey any limitations in our experiment in section~\ref{sec:perspective}. We conclude our paper with final remarks in section~\ref{sec:conclusion}.

\section{Background}\label{sec:background}

\subsection{Kolmogorov-Arnold Networks}

In 2024, Kolmogorov-Arnold Network was proposed by \cite{b9} as an alternative to MLPs. This network was inspired by the Kolmogorov-Arnold representation theorem, stating that any real continuous multivariate function can be represented in terms of a finite number of compositions and addition of continuous univariate functions \cite{b6}. Some conditions of this theorem are relaxed in KAN; namely, splines are used as univariate functions, and the number of summations and compositions is treated as a hyperparameter. Matrices of univariate functions are the key components of KAN. Specifically, let $\Phi_{\ell}$ be an $n_{out}\times n_{in}$ matrix of real-valued 1-D splines. Each such matrix represents a separate layer $\ell$ in the network, where $n_{in}$ is the number of the input variables to the layer and $n_{out}$ is the number of the output variables. These variables passed to the $\ell$-th layer are stored in a vector $\mathbf{x}_\ell$. So the action of one layer can be written in a matrix form in the following way
\begin{align}
    \mathbf{x}_{\ell+1}=\Phi_{\ell}\mathbf{x}_{\ell}.
    \label{eq2}
\end{align}
The action of the whole network consisting of $L$ layers taking $\mathbf{x}$ as an input vector is defined as
\begin{align}
    \mathbf{KAN}(\mathbf{x})=(\Phi_{L-1}\circ\Phi_{L-2}\circ\dots\circ\Phi_{0})\mathbf{x}.
    \label{eq3}
\end{align}
So the two primary characteristics of  KAN are 
\begin{itemize}
\item B-Splines are used as univariate functions. They can easily approximate continuous functions and are differentiable, which is crucial for modern optimization techniques in machine learning. 
\item The number of functions and summations is not restricted to the one shown in the Kolmogorov-Arnold theorem—KAN has an arbitrary number of compositions and summations, which is set by the user. 
\end{itemize}
KAN was originally designed for scientific tasks; therefore, its application to CHILI falls within its intended scope\cite{b9}.

\subsection{Graph-Neural Networks}

Before we begin with the structure of GNNs, it is important to discuss how graphs are represented. A graph is denoted as $G=(V, E)$, where $V$ is the set of its nodes, $E$ is the set of its edges, so $N=|V|$ is the number of nodes and $|E|$ is the number of edges. A graph is conveniently stored in an adjacency matrix $A\in \mathbb{R}^{N\times N}$, where $A_{i,j}$ is $1$ if nodes $i$ and $j$ are connected and $0$ otherwise. Each of the nodes has some features, e.g., if one works with molecules, the nodes are atoms, and their features might be atomic mass, charge, etc. These features are stored in a separate feature matrix $X\in \mathbb{R}^{N\times K}$,  where each row corresponds to a node each having $K$ features. Sometimes the edges also have some features, which are stored in the corresponding edge-feature matrix.

This work focuses on three particular types of GNNs: GCN, GIN, and EdgeCNN. We begin with the \textbf{GCN} introduced by \cite{b8}. Define an adjacency matrix with self-loops $\tilde{A}=A+I_N$, the degree matrix with entries $\tilde{D}_{ii}=\sum_j{\tilde{A}_{ij}}$, and its inverse square root $\tilde{D}_{ii}^{-\frac{1}{2}}=\frac{1}{\sqrt{\tilde{D}_{ii}}}$. Then we can write the equation for a so-called graph convolutional layer as
\begin{align}
    H^{(\ell+1)}=\sigma(\tilde{D}^{-\frac{1}{2}}\tilde{A}\tilde{D}^{-\frac{1}{2}}H^{(\ell)}W^{(\ell)}).
    \label{eq4}
\end{align}
Here $W\in \mathbb{R}^{K \times F}$ is the matrix of learnable weights (where $F$ is the number of neurons in layer $\ell+1$ and $K$ is the number of node features), $\sigma$ is some activation function, and $H$ is the matrix of updated features of nodes (i.e., $H^{(0)}=X$), superscript $(\ell)$ indicates the number of a layer (normally, the layers are stacked one after another to increase efficiency).

Let $v$ be any node with a feature vector $x_v\in \mathbb{R}^K$, which turns into a hidden state vector $h^{(\ell)}_v$ after $\ell$ updates. Then we can rewrite the GCN layer in terms of feature vectors in the following way:
\begin{align}
    h_v^{(\ell+1)} = \sigma  \bigg(W^{(\ell)}\sum_{u\in N(v)\cup\{v\}} \frac{h_u^{(\ell)}}{\sqrt{(deg(v)+1)(deg(u)+1)}} \bigg).
    \label{eq5}
\end{align}
Here, $N(v)$ is a set of neighbors of node $v$ (i.e., nodes directly connected to $v$). The cardinality of $N(v)$ is referred to as the degree of $v$ and is denoted by $deg(v)$. In other words, on each iteration the hidden state of every node is updated based on its neighbors, i.e., the neighboring nodes exchange information. Such an exchange is referred to as message passing.

A GCN layer maps isomorphic graphs (i.e., graphs differing only by the numeration of nodes) to the same outputs. However, this map is not injective in the sense that it might map two non-isomorphic graphs to the same output as well. This may pose a limitation to the representational power of GCNs, and it is not a trivial problem to alleviate it entirely. However, an approach that significantly mitigates this issue---\textbf{GIN}---was introduced by \cite{b17}. It is a type of GNN with discriminative power equivalent to the Weisfeiler–Lehman graph isomorphism test\cite{lehman}. Following the same notation as before, we can describe a GIN layer as
\begin{align}
    h_v^{(\ell+1)} = M_{\ell}\bigg(\big(1+\epsilon^{(\ell)} \big)\cdot h_v^{(\ell)}+\sum_{u\in N(v)} h_u^{(\ell)} \bigg)
    \label{eq6}
\end{align}
where $M_{\ell}$ is an MLP, $\epsilon^{(\ell)}$ is a hyperparameter; in this work we use its default value, zero.
\par
\textbf{EdgeCNN}, proposed by \cite{b16}, is another approach that preserves both the local and the global structure of the graph. The layer consists of two stages. The first one is computing the edges between the nodes, which is described via 
\begin{align}
    e'_{ij\ell} = M_{\ell}\bigg(h_i^{(\ell)}, h_j^{(\ell)}-h_i^{(\ell)}\bigg)
    \label{eq7}
\end{align}
where $e'_{ij\ell}$ is the feature vector of the edge between nodes $i$ and $j$ on the $\ell$-th iteration. The second stage is updating the nodes' representations via a permutation-invariant function like summation or sometimes the max function.
\begin{align}
    h_{i}^{(\ell+1)} = \sum_{j:(i,j)\in E}e'_{ij\ell} 
    \label{eq8}
\end{align}
Having set our notation, we now survey prior KAN-based graph-network methods in Section~\ref{sec:related}.

\section{Related Work}\label{sec:related}

Several studies applied the Kolmogorov-Arnold Representation theorem in machine learning even before \cite{b9}, for instance \cite{b7, b14}. However, these approaches required exactly two layers with fixed number of channels predetermined by the theorem. Once the efficient implementation of KAN suitable for deep learning was released by \cite{b9}, several studies managed to apply it to a number of tasks, including graph-structured data \cite{zhang, b30, b1, b11, b27, b9, Yao_2025}. These studies reviewed the existing graph neural network architectures, such as GCN \cite{b8}, Graph Attention Networks (GAT) \cite{b15}, GraphSAGE\cite{sage} and GIN \cite{b17}, and proposed KAN-based architectures upon them. These new architectures outperformed MLP-based GNNs on the existing graph datasets, including organic molecule datasets MoleculeNet \cite{b31}, ZINC \cite{b28}, and QM9 \cite{b29}. The recent publication of CHILI \cite{b2} introduced the first large-scale publicly available dataset of inorganic nanomaterials, facilitating the development of graph neural networks in this domain. Of course, there are still some GNNs that do not have KAN implementation yet, for instance, Graph Recurrent Networks \cite{b18} and EdgeCNN \cite{b16}. The implementation of the latter is introduced in the next section.

\section{Experiment}\label{sec:method}

\subsection{Models}

The idea to apply KAN to graph-structured data is recent, so there is no consensus on the precise technique. A fairly simple approach is to substitute the activation function in \eqref{eq5} and \eqref{eq6} with a KAN layer \cite{b1}. So the \textbf{KAGCN} layer is defined as follows:
\begin{align}
    h_v^{(\ell+1)} = \mathbf{\Phi_{\ell}}  \bigg(\sum_{u\in N(v)\cup\{v\}} \frac{h_u^{(\ell)}}{\sqrt{(deg(v)+1)(deg(u)+1)}} \bigg).
    \label{eq9}
\end{align}
The \textbf{KAGIN} layer is defined in the following way:
\begin{align}
    h_v^{(\ell+1)} = \mathbf{\Phi_{\ell}}\bigg(\big(1+\epsilon^{(\ell)} \big)\cdot h_v^{(\ell)}+\sum_{u\in N(v)} h_u^{(\ell)} \bigg).
    \label{eq10}
\end{align}

There is no need for the matrix of learnable weights, as KAN does not have them. Therefore, a matrix form of KAGCN is obtained from \eqref{eq4} as follows:
\begin{align}
    H^{(\ell+1)}=\mathbf{\Phi_{\ell}}(\tilde{D}_{ii}^{-\frac{1}{2}}\tilde{A}\tilde{D}_{ii}^{-\frac{1}{2}}H^{(\ell)}).
    \label{eq11}
\end{align}
To the best of our knowledge there is no implementation of EdgeCNN based on KAN (KAEdgeCNN), so we conclude this section by introducing the \textbf{KAEdgeCNN} layer in a similar fashion as \eqref{eq7} and \eqref{eq8}, substituting MLP with a KAN.
\begin{align}
    h_{i}^{(\ell+1)} = \sum_{j:(i,j)\in E}\mathbf{\Phi_{\ell}}\bigg(h_i^{(\ell)}, h_j^{(\ell)}-h_i^{(\ell)} \bigg)
    \label{eq12}
\end{align}
The architecture of the models using the aforedescribed layers is elaborated in appendix~\ref{app:model_arch}.

\subsection{Dataset and Tasks}

CHILI poses six material design tasks and eleven property prediction tasks \cite{b2}. It consists of two branches of datasets CHILI-3k and CHILI-100k labeled by their size. The authors of CHILI benchmarked seven popular GNNs (three of which are GCN, GIN, and EdgeCNN) on eight property prediction tasks: atomic number classification, crystal system classification, space group classification, absolute position regression, edge-attribute (distance) regression, SAXS (small-angle X-ray scattering) regression, XRD (X-ray diffraction) regression, xPDF (X-ray pair-distribution function) regression. (Also, see \cite{osti_10300745} for a context). The classification tasks are evaluated using the weighted F1-score, absolute position regression is evaluated using the mean absolute error (MAE), and the remaining regression tasks are evaluated using the mean squared error (MSE). We select the same eight tasks for a comparison between the GNN and KAGNN models in our experiment using the same metrics. The technical details of the usage of the data are described in Appendix~\ref{app:dataset}.

\begin{table*}[htbp]
\setlength{\tabcolsep}{1mm}
\begin{center}
\begin{tabular}{|c|c|c|c|c|c|c|c|}
\hline

\textbf{Task (metrics)}
  & \textbf{KAGCN}
  & \textbf{KAGIN}
  & \textbf{KAEdgeCNN}
  & \textbf{GCN}
  & \textbf{GIN}
  & \textbf{EdgeCNN}
  & \textbf{Metrics}\\
\hline
\multicolumn{8}{|c|}{\textbf{CHILI-3K}} \\
\hline
edge attr.     & \textbf{0.050$\pm$0.007} & \textbf{0.058$\pm$0.009} & $0.031\pm0.014$ & $0.056\pm0.006$ & $0.464\pm0.005$ & $0.015\pm0.001$ & MSE \\
cryst. sys.  & \underline{\textbf{0.995$\pm$0.004}} & \textbf{0.671$\pm$0.076} & \textbf{0.976$\pm$0.021} & $0.367\pm0.127$ & $0.438\pm0.004$ & $0.657\pm0.196$ & F1  \\
space grp.      & \textbf{0.937$\pm$0.075} & \textbf{0.547$\pm$0.137} & \underline{\textbf{0.966$\pm$0.011}} & $0.099\pm0.019$ & $0.125\pm0.026$ & $0.733\pm0.207$ & F1\\
saxs               & $0.036\pm0.001$ & $0.037\pm0.000$ & $1.545\pm2.133$   & $0.008\pm0.000$ & $0.008\pm0.000$ & $0.006\pm0.004$ & MSE \\
xrd                & \underline{\textbf{0.007$\pm$0.000}} & Unstable        & $0.157\pm0.179$ & $0.010\pm0.000$ & Unstable        & $0.008\pm0.001$ & MSE \\
xPDF             & \textbf{0.010$\pm$0.000} & \textbf{0.070$\pm$0.001} & $0.037\pm0.021$            & $0.012\pm0.000$ & Unstable        & $0.011\pm0.000$ &  MSE \\
abs. pos        & $16.698\pm0.000$ & $16.698\pm0.000$ & 16.698$\pm$0.000            & $16.575\pm0.000$ & $16.575\pm0.000$ & $16.575\pm0.000$ &  MAE  \\
atom cls.          & \textbf{0.627$\pm$0.001} & \textbf{0.613$\pm$0.005} & \textbf{0.657$\pm$0.003}            & $0.496\pm0.001$ & $0.587\pm0.002$ & $0.632\pm0.009$ & F1  \\
\hline
\multicolumn{8}{|c|}{\textbf{CHILI-100K}} \\
\hline
edge attr.        & $0.098\pm0.019$ & \textbf{0.110$\pm$0.015} & 0.040$\pm$0.004            & $0.090\pm0.002$ & $0.491\pm0.038$ & $0.030\pm0.001$  & MSE       \\
cryst. sys.  & \underline{\textbf{0.477$\pm$0.008}} & \textbf{0.190$\pm$0.034} & \textbf{0.295$\pm$0.024}            & $0.069\pm0.023$ & $0.069\pm0.040$ & $0.072\pm0.047$ & F1 \\
space grp.     & \underline{\textbf{0.216$\pm$0.009}} & \textbf{0.052$\pm$0.009} & 0.118$\pm$0.012            & $0.043\pm0.001$ & $0.043\pm0.000$ & $0.158\pm0.035$ & F1\\
saxs                 & $0.0378\pm0.000$ & Unstable        & 0.089$\pm$0.070            & $0.010\pm0.000$ & $0.009\pm0.000$ & $0.007\pm0.009$ & MSE        \\
xrd                  & \textbf{0.007$\pm$0.000} & Unstable        & 0.084$\pm$0.105            & $0.009\pm0.000$ & $0.009\pm0.000$ & $0.006\pm0.000$  & MSE   \\
xPDF                  & \textbf{0.011$\pm$0.000} & $0.089\pm0.006$ & 0.452$\pm$0.613            & $0.014\pm0.000$ & $0.013\pm0.000$ & $0.012\pm0.000$ & MSE        \\
abs. pos        & \textbf{16.182$\pm$0.000} & \textbf{16.182$\pm$0.000} & \textbf{16.182$\pm$0.000}            & $16.336\pm0.000$ & $16.336\pm0.000$ & $16.336\pm0.000$ & MAE         \\
atom cls.       & \textbf{0.317$\pm$0.002}  & \textbf{0.374$\pm$0.002} & 0.449$\pm$0.002            & $0.275\pm0.002$ & $0.336\pm0.005$ & $0.572\pm0.017$  & F1 \\
\hline
\end{tabular}

\caption{The results of the KAGCN, KAGIN, and KAEdgeCNN models on the CHILI datasets.}
\label{tab:chili3k_results}
\end{center}
\end{table*}

\subsection{Method}\label{sec:experiments}

To construct the three proposed models, KAGCN, KAGIN and KAEdgeCNN, we used PyTorch \cite{b19}. All KAN models here are based on a publicly available efficient-KAN\footnote{https://github.com/Blealtan/efficient-kan} implementation. The models were trained and evaluated on an NVIDIA GeForce RTX 3080 GPU with 10 GB of GDDR6X memory. First, we conducted a hyperparameter search on CHILI-3K, the result of which we state in Appendix~\ref{app:hyperparam}. To measure the performance, we trained each of the models with the found hyperparameters three times with different seeds. Each model was trained for 500 epochs with early stopping and a patience of 20 epochs (i.e., training was terminated if the validation accuracy did not improve for 20 consecutive epochs). The mean, standard deviation, and metrics of the results are reported in Table~\ref{tab:chili3k_results} (we used the same metrics as \cite{b2}). The label ``Unstable" was assigned to the cases where the performance distributions were dispersed, indicating unreliable convergence. In the table, we also provide the performance of vanilla GCN, GIN, and EdgeCNN taken from \cite{b2}. We mark in bold the metrics of KAN models that outperform the corresponding MLP model on a particular task. State-of-the-art results across all models (including those beyond GCN, GIN and EdgeCNN) are underlined.

\subsection{Results and Interpretations}

We find that KAGCN and KAGIN improved in performance (compared to their respective counterparts) on all classification tasks by a large margin, and KAEdgeCNN improved on all classification tasks of CHILI-3K and one classification task from CHILI-100K. In particular, KAGCN significantly outperformed all previous results on crystal system classification, achieving a weighted F1-score of 0.995, whereas the best result from the non-KAN models was 0.657. We exclude the possibility of data-leaking, as the test set was never used during the hyperparameter tuning and training stages. Whilst the classification performance has been improved, the results on regression tasks remain at a comparable level to those of standard GNNs. We hypothesize it is due to the disproportionately higher difficulty of the regression tasks in the CHILI datasets compared to the classification ones, which is stated by the authors of the dataset as well.

Some of the observed performance patterns are closely linked to the structural properties of the dataset. Generally higher performances on the crystal system classification task compared to the space group classification task are likely due to the significant difference in the number of classes: 7 and 230 respectively. Nonetheless, KAGCN and KAEdgeCNN achieve F1 scores above 0.9 on both tasks on CHILI-3K. The difficulty gap between the two versions of the dataset lies not just in size but in diversity. While CHILI-3K contains only four crystal systems, CHILI-100K contains seven. In CHILI-3K, each nanomaterial consists of 2 elements, while in CHILI-100k this number reaches 7. 

EdgeCNN usually performs better on the CHILI dataset than any other MLP-based GNNs; hence, it was expected that KAEdgeCNN would analogously outperform all KAGNNs. However, for KAN-based GNNs, these differences in results have been largely diminished, or even reversed in some cases. On CHILI-3K, there is little difference in performance between KAGCN and KAEdgeCNN, and on CHILI-100k, KAGCN outperforms KAEdgeCNN on several tasks. It is plausible that the incorporation of KAN layers lead to the detection of more patterns that previous GNN architectures could not fully take advantage of.

We claim three new state-of-the-art results on the CHILI-3K dataset and two on CHILI-100k by KAGNN models. For CHILI-3k, KAGCN performs best on the crystal system classification, KAEdgeCNN on the space group classification, and KAGCN on the XRD regression. For CHILI-100k, the best performances on both crystal system and space group classifications are claimed by KAGCN.

\section{Limitations}\label{sec:perspective}

High performance of KAGNN comes at the cost of a large number of learnable parameters. While the number of parameters in the vanilla GNN models, except for the EdgeCNN, ranges between 2 and 8 thousand, KAGNNs in our case exhibit substantially larger parameter counts (see Table~\ref{tab:trainable_params}). This difference is due to the utilization of splines, the implementation of which requires the use of a higher number of learnable weights. Nonetheless, the marginal efficiency of the trainable parameters might be worth their quantity. For instance, in the crystal system classification task, KAGCN outperformed EdgeCNN, while the latter had almost twice as many parameters as the former. Thus, it is yet to be explored when it is reasonable to use KAGNNs instead of the vanilla GNN to make efficient use of memory resources.

It was impossible for us to train KAEdgeCNN with more than one hidden layer due to its size: EdgeCNN already had over 40000 trainable parameters \cite{b2}. Together with splines in KAN taking up large memories, it is likely we did not explore the model's full potential. It might be worthwhile implementing this model with a more lightweight kernel, e.g., an RBF-based KAN implementation, which is also publicly available\footnote{https://github.com/ZiyaoLi/fast-kan}, or Fourier-based KAN as in\cite{b11, fourierkan}.

It is evident that the regression tasks in both CHILI-3K and CHILI-100K remain challenging for all modifications of KAGNNs; thus, these tasks require a further investigation with a different approach.

In this study, we neglected the existing KAN implementations of GAT which also could have been a good comparison \cite{b1, b11}.




\section{Conclusion}\label{sec:conclusion}

In this work, we extended the family of KAN-based GNN models by introducing KAEdgeCNN and investigated the potential of KAGNNs on inorganic nanomaterials data using KAGCN, KAGIN, and the newly developed KAEdgeCNN. For this, we conducted a hyperparameter search and made publicly available both the parameters and the trained models. We conducted a comparative analysis between MLP- and KAN-based GNNs and reported significant improvements on the classification tasks. However, these improvements often required an increase in the number of trainable parameters. We also observed that applying KAN instead of MLP eliminates the performance disparity between different types of GNNs, bringing their performance to a comparable level. In conclusion, we state that KAGNNs show high potentials not only on organic-molecule datasets as demonstrated in the previous studies but also on datasets of inorganic nanomaterials.

\bibliographystyle{IEEEtran}
\bibliography{bibliography}

\appendix
\subsection{Hyperparameter Search}\label{app:hyperparam}
The hyperparameter search was conducted only on CHILI-3K. We did not conduct a separate tuning for the CHILI-100K version to spare the resources. We used Optuna \cite{optuna} for hyperparameter tuning with 40 trials. The training was conducted for 500 epochs with patience 20. The batch size was set to 32, as larger batch sizes could potentially lead to out-of-memory errors. We share the obtained  hyperparameters in Table~\ref{tab:best_hyperparameters} and the search range for them in Table~\ref{tab:search_range}.

\begin{table}[htbp]
\setlength{\tabcolsep}{1mm}
\begin{center}
\begin{tabular}{|c|c|c|c|c|c|}
\hline
\textbf{Learning rate} & \textbf{Layers} & \textbf{Hidden dim.}& \textbf{Dropout} & \textbf{Grid size} & \textbf{Spline ord.} \\
\hline
$[10^{-4}, 10^{-2}]$        & $[1, 3]$ & $[16, 64]$ & $[0, 0.5]$&$[3, 5]$&$[3, 5]$\\
\hline
\end{tabular}

\caption{The search range for the hyperparameters.}
\label{tab:search_range}
\end{center}
\end{table}

\begin{table}[htbp]
\setlength{\tabcolsep}{1mm}
\begin{center}

\begin{tabular}{|c|c|c|c|c|c|c|}
\hline
\textbf{Task} & \textbf{L.r.} & \textbf{Layers} & \textbf{H.dim.} & \textbf{Dropout} & \textbf{Grid} & \textbf{Spline} \\
\hline
\multicolumn{7}{|c|}{\textbf{KAGCN}} \\
\hline
edge attr.       & 0.0023 & 3 & 24 & 0.0247 & 4 & 3 \\
cryst.\ sys. & 0.0069 & 2 & 39 & 0.2141 & 4 & 4 \\
space grp.     & 0.0045 & 3 & 59 & 0.0711 & 3 & 5 \\
saxs                & 0.0093 & 1 & 50 & 0.3521 & 3 & 4 \\
xrd                    & 0.0005 & 1 & 48 & 0.4510 & 5 & 3 \\
xPDF              & 0.0015 & 1 & 33 & 0.3182 & 4 & 4 \\
abs.\ pos.      & 0.0007 & 2 & 37 & 0.4668 & 5 & 4 \\
atom cls.              & 0.0001 & 3 & 55 & 0.1424 & 5 & 3 \\
\hline
\multicolumn{7}{|c|}{\textbf{KAGIN}} \\
\hline
edge attr.       & 0.0055 & 3 & 36 & 0.0618 & 3 & 3 \\
cryst.\ sys.    & 0.0011 & 1 & 64 & 0.3012 & 5 & 4 \\
space grp.    & 0.0012 & 2 & 64 & 0.2179 & 5 & 3 \\
saxs             & 0.0054 & 1 & 34 & 0.2496 & 3 & 5 \\
xrd                  & 0.0011 & 1 & 33 & 0.1910 & 3 & 4 \\
xPDF                  & 0.0013 & 1 & 34 & 0.4379 & 5 & 4 \\
abs.\ pos.      & 0.0005 & 1 & 53 & 0.4079 & 5 & 3 \\
atom cls.            & 0.0009 & 3 & 19 & 0.1436 & 5 & 3 \\
\hline
\multicolumn{7}{|c|}{\textbf{KAEdge}} \\
\hline
edge attr.       & 0.0017 & 1 & 35 & 0.0217 & 4 & 4 \\
cryst.\ sys.    & 0.0042 & 1 & 42 & 0.0407 & 3 & 3 \\
space grp.    & 0.0028 & 1 & 63 & 0.4753 & 3 & 5 \\
saxs             & 0.0052 & 1 & 39 & 0.3192 & 3 & 5 \\
xrd                  & 0.0005 & 1 & 64 & 0.4767 & 4 & 5 \\
xPDF                  & 0.0020 & 1 & 54 & 0.0865 & 3 & 5 \\
abs.\ pos.      & 0.0029 & 1 & 25 & 0.1946 & 3 & 3 \\
atom cls.            & 0.0033 & 1 & 48 & 0.0708 & 5 & 5 \\
\hline
\end{tabular}

\caption{The best hyperparameters found for the KAGNN models.}
\label{tab:best_hyperparameters}
\end{center}
\end{table}

\begin{table}[htbp]
\setlength{\tabcolsep}{1mm}
\begin{center}
\begin{tabular}{|c|c|c|c|}
\hline
\multicolumn{4}{|c|}{\textbf{Trainable Parameters}} \\
\hline
\textbf{Task} & \textbf{KAGCN} & \textbf{KAGIN}& \textbf{KAEdgeCNN} \\
\hline
edge attribute reg.        & 145{,}416 & 204{,}480& 214,970\\
crystal system classif.    &  20{,}748 &   9{,}984& 7,140\\
space group classif.       & 209{,}627 & 192{,}896& 153,846\\
saxs reg.                  &  23{,}850 &  12{,}648& 6,318\\
xrd reg.                   & 577{,}968 & 354{,}717& 826,624\\
xPDF reg.                  & 3{,}970{,}098 & 4{,}501{,}124& 6,487,668\\
absolute position reg.     &  27{,}343 &  31{,}747& 2,850\\
atom classif.              & 188{,}815 &  51{,}604& 136,032\\
\hline
\end{tabular}

\caption{The number of trainable parameters in the KAGNN models.}
\label{tab:trainable_params}
\end{center}
\end{table}

\subsection{Dataset}\label{app:dataset}
The CHILI-3K dataset was used entirely, while in CHILI-100K we selected a subset consisting of 2975 instances stratified on crystal system number, exactly as \cite{b2} did. Regardless of the dataset configuration, we split the data into the training (80\%), validation (10\%), and test (10\%) sets. During the hyperparameter tuning, the accuracy was evaluated on the validation set. The test set was used only during the models' evaluation to measure their final performance.

\subsection{Architecture of the models}\label{app:model_arch}
For KAGCN and KAGIN, the architectures described by \cite{b1} are adopted. Each model is composed of a stack of hidden blocks, followed by a task-dependent head. The structure of a hidden block of each model is depicted in Fig. ~\ref{scheme}. There are three possible task-dependent heads:
\begin{itemize}
    \item KAN-readout layer (for node level tasks);
    \item global average pooling (for graph level tasks);
    \item the dot product between the output vectors of the neighboring nodes (for edge level tasks).
\end{itemize}
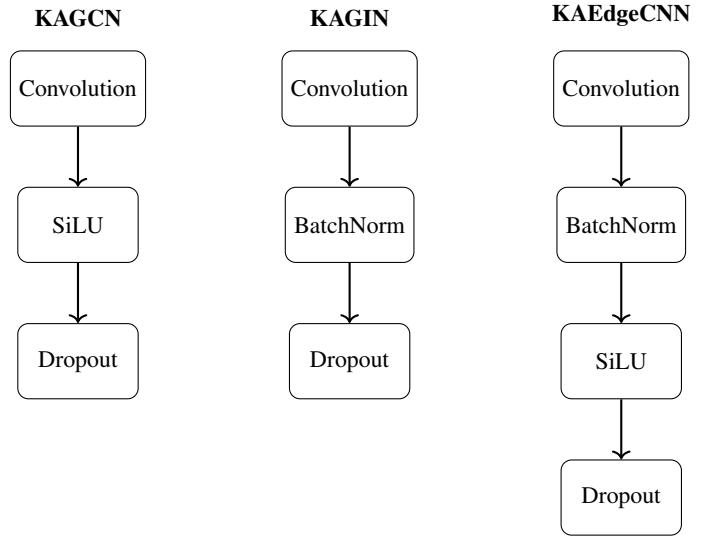
\begin{figure}[t]
\centering
\begin{tikzpicture}[node distance=0.8cm, column sep=1.5cm]

\node[block] (gcn1) {Convolution};
\node[block, below=of gcn1] (gcn2) {SiLU};
\node[block, below=of gcn2] (gcn3) {Dropout};

\draw[arrow] (gcn1) -- (gcn2);
\draw[arrow] (gcn2) -- (gcn3);

\node[above=0.2cm of gcn1] {\small\textbf{KAGCN}};

\node[block, right=1.8cm of gcn1] (gin1) {Convolution};
\node[block, below=of gin1] (gin2) {BatchNorm};
\node[block, below=of gin2] (gin3) {Dropout};

\draw[arrow] (gin1) -- (gin2);
\draw[arrow] (gin2) -- (gin3);

\node[above=0.2cm of gin1] {\small\textbf{KAGIN}};

\node[block, right=1.8cm of gin1] (edge1) {Convolution};
\node[block, below=of edge1] (edge2) {BatchNorm};
\node[block, below=of edge2] (edge3) {SiLU};
\node[block, below=of edge3] (edge4) {Dropout};

\draw[arrow] (edge1) -- (edge2);
\draw[arrow] (edge2) -- (edge3);
\draw[arrow] (edge3) -- (edge4);

\node[above=0.2cm of edge1] {\small\textbf{KAEdgeCNN}};

\end{tikzpicture}
\caption{Hidden block structure.}
\label{scheme}
\end{figure}

\end{document}